\documentclass[letterpaper]{article}
\usepackage{iccc}
\usepackage{times}
\usepackage{helvet}
\usepackage{courier}
\usepackage{tcolorbox}
\usepackage{tabularx}

\title{Solving and Generating NPR Sunday Puzzles with Large Language Models}
\author{Jingmiao Zhao and Carolyn Jane Anderson\\
Computer Science Department\\
Wellesley College\\
Wellesley, MA 02482 USA\\
carolyn.anderson@wellesley.edu\\
}
\begin{document} 
\maketitle
\begin{abstract}
We explore the ability of large language models to solve and generate puzzles from the NPR Sunday Puzzle game show using PUZZLEQA, a dataset comprising 15 years of on-air puzzles. We evaluate four large language models using PUZZLEQA, in both multiple choice and free response formats, and explore two prompt engineering techniques to improve free response performance: chain-of-thought reasoning and prompt summarization. We find that state-of-the-art large language models can solve many PUZZLEQA puzzles: the best model, GPT-3.5, achieves 50.2\% loose accuracy. However, in our few-shot puzzle generation experiment, we find no evidence that models can generate puzzles: GPT-3.5 generates puzzles with answers that do not conform to the generated rules. Puzzle generation remains a challenging task for future work. 
\end{abstract}

\section{Introduction}

Puzzles and games have long been used to benchmark progress in AI. We continue this tradition by exploring the ability of large language models (LLMs) to solve word puzzles from the NPR Sunday Puzzle on-air game show. Recent advances have led to new techniques for using general-purpose text generation models to solve a variety of tasks. In few-shot learning, a model is prompted with a handful of examples and asked to generate a solution. In prompt engineering, the input to the model is manipulated in order to improve the model's performance on a task. These techniques have led to surprisingly good performance by LLMs on novel tasks, without any further training of the model. 

In this paper, we explore whether few-shot learning and prompt engineering can allow LLMs to solve questions from the NPR Sunday Puzzle game show, which combines information retrieval, wordplay, and pattern recognition. We introduce PUZZLEQA, consisting of puzzle descriptions, questions, and answers for 558 puzzles, and use it to benchmark four state-of-the-art LLMs. We explore prompt engineering techniques, but find that they have little impact on performance. We also explore whether models can generate new puzzles and find that this remains a challenging task. Although the best model, GPT-3.5, is capable of solving 50.2\% of the puzzles, it cannot generate playable games.

\section{Benchmarking AI through Games}

Our work continues the tradition of evaluating AI progress through puzzles and games~\cite{6177724,rodriguez2021quizbowl,rozner_decrypting_2021,sobieszek_playing_2022}. Contemporary LLMs have demonstrated strong performance on a wide variety of language tasks, including question-answering. However, the extent of their ability to generalize patterns and to solve wordplay puzzles is under-explored. 

The NPR Sunday Puzzle game show represents a particularly interesting genre of puzzle to explore because it synthesizes a variety of skills: information retrieval; rhyming, anagram-solving, and other wordplay; and pattern recognition. Figure \ref{fig:ex_puzzle} shows an example of a puzzle, which involves knowledge of country names and wordplay. Despite the complexity of some NPR Sunday Puzzle games, compared to other question-answering games used to benchmark LLMs, such as Jeopardy! and Quiz Bowl, they are targeted towards a broader audience and require less specialized knowledge.

\begin{figure}
    \begin{tcolorbox}
    \textbf{Puzzle Description}: Today's puzzle involves ``consonyms,'' which are words that have the same consonants in the same order but with different vowels. Every answer is the name of a country.\\
    \textbf{Question}: MINGLE\\
    \textbf{Answer}: MONGOLIA
    \end{tcolorbox}
    \caption{NPR Sunday Puzzle from March 12, 2023}  \label{fig:ex_puzzle}
\end{figure}

\begin{figure*}[t]
    \centering
    \includegraphics[width=\textwidth]{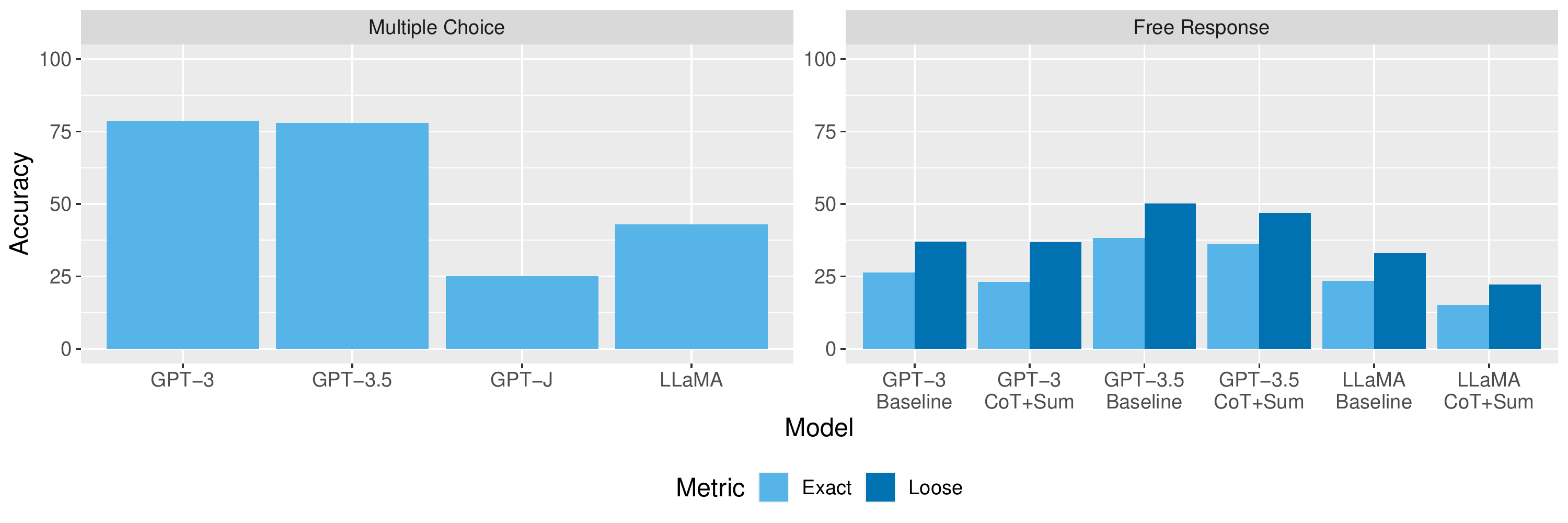}
    \caption{Results on full PUZZLEQA dataset, by model, prompting technique, and format.}
    \label{fig:bigexperiment}
\end{figure*}

\subsection{Dataset}

We present PUZZLEQA, a dataset of 558 NPR Sunday Puzzle games from 2007-2021. During this period, a group of fans ran a mailing list, NPR Puzzle Synopsis, that distributed questions and answers for each week's puzzle.\footnote{https://groups.google.com/g/nprpuzzle} We obtained the puzzle explanations from the NPR website,\footnote{https://www.npr.org/series/4473090/sunday-puzzle.} and extracted the answers from the mailing list, using GPT-J to aid in preprocessing the data.  We also classified the puzzles into 11 different categories. The dataset, preprocessing tools, and analysis scripts will be released publicly.\footnote{https://github.com/Wellesley-EASEL-lab/PuzzleQA}

\subsection{Model Selection}

We explored two publicly available LLMs, GPT-J~\cite{gpt-j} and LLaMA~\cite{touvron2023llama}, and two proprietary OpenAI LLMs: GPT-3 Davinci and GPT-3.5~\cite{gpt3}. The amount of randomness in each of these model's generations can be manipulated via the temperature hyperparameter, where a high temperature means more randomness. After exploring temperature settings of 0.75, 0.5, 0.25, and 0.1, we found that temperature = 0.1 was optimal. 

\section{Multiple Choice Experiments}

As an easier benchmark, we constructed a multiple-choice version of the PUZZLEQA dataset. For each problem, we randomly selected three answers to other questions from the same puzzle to present alongside the correct answer. 

\paragraph{Answer only baseline}

In multiple choice tasks, there can be biases towards or against certain question options, even in the absence of the question. To obtain an accurate baseline, we measure how often the model selects the correct answer when it is not given the question. An unbiased set of answer options would result in at-chance performance (25\%). We refer to this task as the answer-only baseline. We find that the model selects the correct answer 21\% of the time when it is not given the question, suggesting that there is no significant bias towards the correct answer from the answer options alone.

\paragraph{Results} 
Figure \ref{fig:bigexperiment} shows the performance of each model on the multiple choice task. The smallest model, GPT-J, does not perform better than chance on this simplified task. As a result, we exclude it from the rest of our experiments. The other publicly available model, LLaMA, performs well above chance, showing that it is able to correctly identify responses for many problems. GPT-3 and GPT-3.5 both perform well on this task, solving 78\% percent of problems.

\section{Free Response Experiments}

We perform two sets of free response experiments. To explore various prompt engineering techniques, we first create a subset of our data balanced by question type. We then compare the best prompting technique against a baseline on the full dataset. In all experiments, a few-shot paradigm is used: the model is given two examples of solved questions from the same puzzle (following the same game rules) and asked to solve a third.

\subsection{Evaluation Metrics}

Free response question-answering is difficult to evaluate, since a correct answer may be phrased in various ways. We use two conservative metrics for evaluating performance.

\textit{Exact Matching}: the response is correct if it exactly matches the gold solution.

\textit{Loose Matching}: the response is correct if it is contained or contains the gold solution, after removing non-alphabetical characters and lowercasing both strings.

\subsection{Exploring Prompt Engineering Techniques}

We subsample our dataset in order to evaluate the impact of various prompt engineering techniques. 10 questions from each of our 11 categories were randomly sampled for the subset, for a total of 110 items. We explore two prompt engineering techniques: summarization and chain-of-thought reasoning.

\subsubsection{Summarization}

One potential challenge for the model in solving the PUZZLEQA puzzles is that the games are described informally. We hypothesize that the lack of consistency in puzzle wording might hinder the model. We experiment with using GPT-3.5 to summarize the puzzle description to a more consistent format (Figure \ref{fig:summarization}).

\begin{figure}[t]
\begin{tcolorbox}
\textbf{Summarize the following:}
In the on-air puzzle, you are given the word and must drop two letters so that the remaining letters, in order, spell a color or shade.
\end{tcolorbox}
\caption{Summarization prompt to summarized explanations of the rules of the puzzle}\label{fig:summarization}
\end{figure}

\subsubsection{Chain-of-thought Reasoning}

Prompting models to explain their reasoning before generating an answer has been shown to improve model performance on other tasks~\cite{wei2023chainofthought}. This is known as \textit{chain-of-thought prompting}. 

One limitation of this approach is that humans must write explanations to provide as examples to the model. We automate the process by using the model to generate explanations for rule-question-answer triplets. We then use the generated explanations as input to the chain-of-thought prompting experiment. Figure \ref{fig:reasoning} shows an example prompt used to gather model explanations. GPT-3's generated explanation was \textit{The word ``blouse'' can have two letters dropped to spell the color ``blue''}. This explanation was then added to the example to use in few-shot prompting.

\begin{figure}[t]
\begin{tcolorbox}
\textbf{Puzzle Description}:  In the on-air puzzle, you are given the word and must drop two letters so that the remaining letters, in order, spell a color or shade.\\
\textbf{Question}: blouse \\
\textbf{Answer}: blue\\
\textbf{Please explain this answer.}
\end{tcolorbox}
\caption{Chain-of-thought prompt to elicit explanations}\label{fig:reasoning}
\end{figure}

\subsection{Prompt Engineering Results}

\begin{figure}
    \centering
    \includegraphics[width=0.5\textwidth]{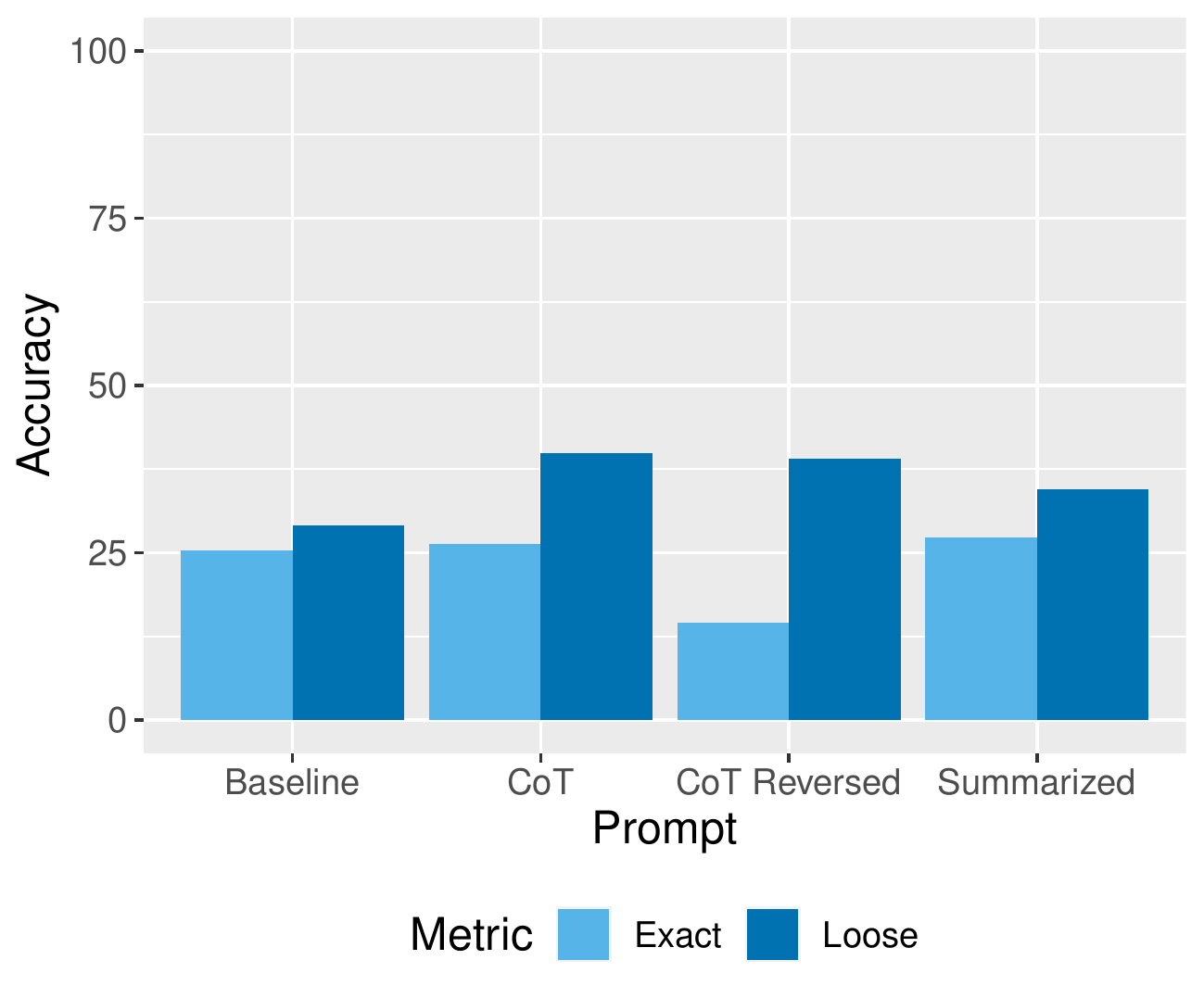}
    \caption{GPT-3 results for prompt engineering experiment}
    \label{fig:small_res}
\end{figure}

In our small-scale experiment, we found that both summarization and chain-of-thought prompting improved performance. Figure \ref{fig:small_res} shows GPT-3 results for each technique.  

\subsection{Free Response Results}

We select the best-performing prompt engineering technique to compare against a baseline prompt on the full PUZZLEQA dataset. Our small-scale experiments suggested that both summarization and chain-of-thought prompting improve performance. We compare this model to a baseline few-shot learning model.

Figure \ref{fig:bigexperiment} shows the free response results for the full PUZZLEQA dataset. Although chain-of-thought reasoning and summarization improved model performance in our small-scale experiment, this did not replicate for the entire dataset. The baseline GPT-3.5 model performs best, solving 50.2\% of the puzzles. We note that performance is very sensitive to prompt wording: when we rephrase the chain-of-thought prompt to ask for the ``answer and reasoning'' rather than the ``reasoning and answer,'' performance drops substantially.

\subsubsection{Discussion}

Overall, the best model, GPT-3.5, performs fairly well on the PUZZLEQA dataset. We observe a large gap in performance between the LLaMA model, which is available for academic research, and the proprietary OpenAI models, illustrating the need for better public LLMs. Surprisingly, we find that the prompt engineering techniques we explored did not improve model performance. 

Although chain-of-thought prompting did not improve performance, we feel that it still has some benefits. The chain-of-thought-prompted model: when prompted this way, GPT-3.5 produces ``N/A'' 87 times, compared to only 3 refusals to provide an answer in the baseline version. In some cases, the puzzle is faulty due to webscraping errors. When we manually examined the explanations generated by the model, most are consistent with the answer. Thus, chain-of-thought prompting may decrease overconfidence in models while providing a window into the model's decisions.

\section{Game Generation}

Although our exploration of popular prompt engineering techniques was not fruitful, we nonetheless found that state-of-the-art LLMs are capable of solving many of the NPR puzzles. In this section, we explore whether LLMs are also capable of generating puzzles for humans to play. We explore puzzle generation with the LLM that achieved the highest performance on the free response task, GPT-3.5.

\subsection{Prompt Design}

We construct a few-shot puzzle generation dataset using our balanced 110 question subset. In each prompt, we provide the model with five examples of puzzles, presented as explanation-question-answer triplets (Figure \ref{fig:game_gen}), for a total of 22 prompts. We then evaluate each generated puzzle by asking the model to generate an answer to the question.

\begin{figure}
\begin{tcolorbox}
You are given several examples of the game, with each game including a prompt, question, and answer. \\
\textit{5 examples given as:\\
Explanation:\\
Question:\\
Answer:\\}
Please generate a new game with a prompt, question, and answer in the same format.
\end{tcolorbox}
\caption{Game generation prompt}\label{fig:game_gen}
\end{figure}

\begin{table*}[t]
  \noindent
  \begin{tabularx}{\linewidth}{|X|c|c|c|}
    \hline
    Explanation & Question & Response & Gold \\\hline
    I'll give you a word that can be split into two smaller words. The first word is the name of a U.S. state, and the second word is a type of animal. For example, given "MontanaLion," the answer would be "Montana, Lion." & DelawareHawk & Delaware, Hawk & Delaware, Hawk\\\hline
    Every answer is a well-known movie title with one or more letters from the title replaced with a number. For example, if the clue is "Th3 Matrix," the answer would be "The Matrix."	& F1ght Club & Fight Club & Fight Club\\\hline
 \end{tabularx} 
 \caption{Game that satisfies both consistency and conformity, but are trivial}
  \label{tab:gamegeneasy}
\end{table*}

\subsection{Evaluation Metrics}

We use two metrics to evaluate the generated games:

\textit{Consistency}: can the model solve its own puzzle? We provide the generated explanation and question to GPT-3.5 and generate an answer. If the answers match, the puzzle is consistent.

\textit{Conformity}: of the questions that are consistent, how many have answers that conform to the rules in the explanation? We assess this manually.

\subsection{Results}

Of the 22 games generated by GPT-3.5, it answers 17 questions consistently. However, just 6 of the questions conform with the explanation provided. In addition, the conforming games are trivial to solve (Table \ref{tab:gamegeneasy}). Thus, though LLMs succeed in playing the NPR Sunday Puzzle, we find no evidence that they can generate new puzzles for human players.

\section{Limitations and Future Work}

\begin{figure}
    \centering
    \includegraphics[width=0.45\textwidth]{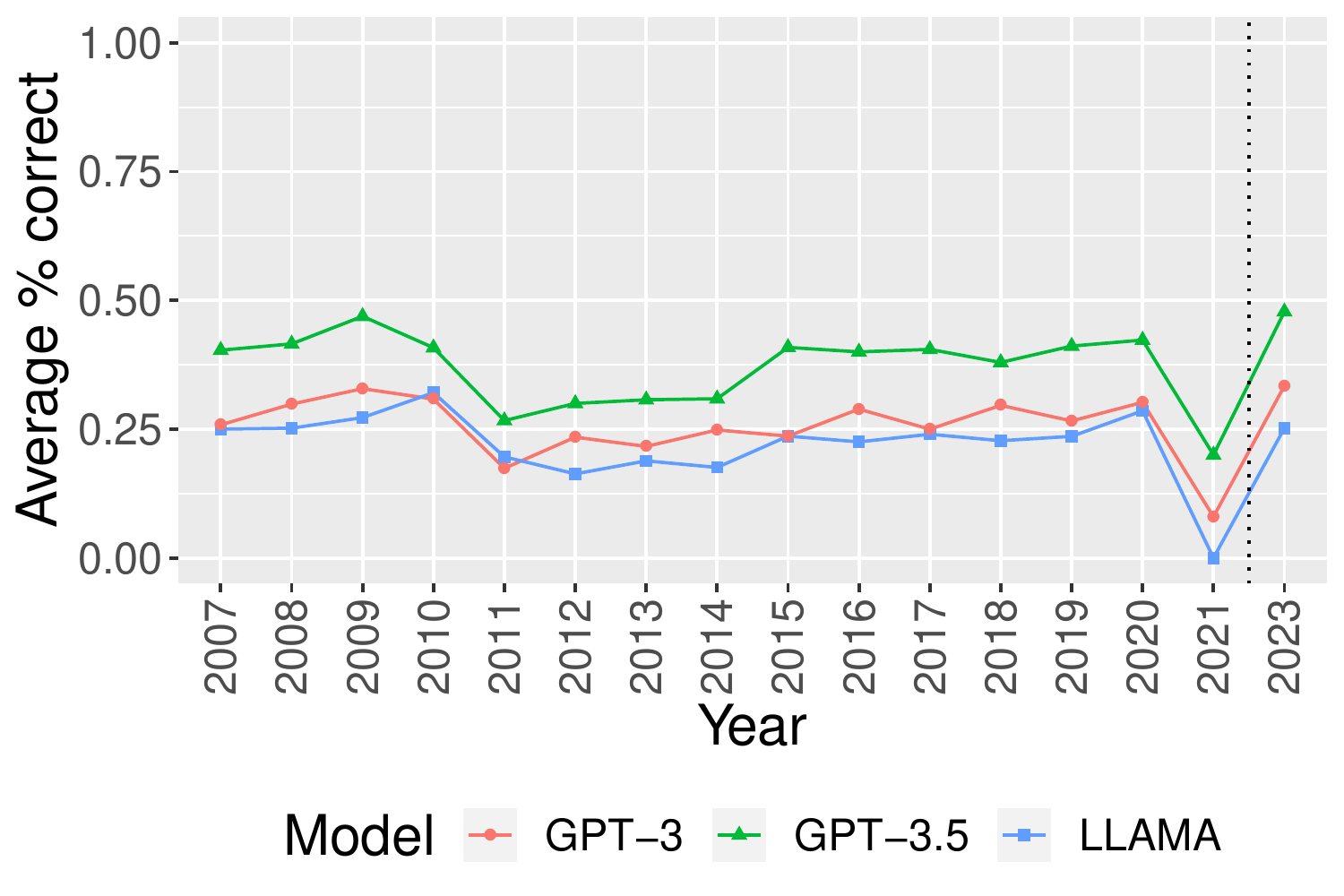}
    \caption{Exact accuracy by puzzle air date}
    \label{fig:time}
\end{figure}

Our experiments with PUZZLEQA show that current LLMs are capable of solving, but not creating, NPR Sunday Puzzle questions. However, our results come with a number of caveats. First, since the training data for GPT-3, GPT-3.5, and LLaMA is not publicly available, we cannot measure whether models have been trained on problems within our dataset. To investigate potential training/test overlap, we manually constructed a test set of questions from 2023, which is more recent than the models' training data cutoff dates (GPT-3: 2019; GPT-3.5: 2021; LLaMA: 2022). We find that model performance on this small (n=116) subset is on par with the full dataset (Figure \ref{fig:time}). In general, although performance varies by year, there is no clear trend.\footnote{We note that puzzle types and topics may vary over time; an in-depth analysis of the puzzle content is one area for future work.}

Our methodology could also be refined in a number of ways. Our webscraping techniques failed to capture some questions, which could be added to our dataset. Our loose accuracy metric is a conservative measure of model capability, since it may fail to identify some valid answers. Finally, future work could incorporate a rating of question difficulty by identifying from the game transcript whether the human player succeeded or failed in answering the question.

\section{Conclusion}

Using data from the NPR Sunday Puzzle game show, we explore the ability of contemporary large language models to solve and generate word puzzles. We show that PUZZLEQA is a challenging benchmark for LLMs: although GPT-3.5 solves 50.2\% of the problems in the free response task, information about its training data is not public, and the best publicly available model achieves only 33\%. 

The fact that the prompt engineering techniques we explored failed to improve performance is puzzling, given promising results from chain-of-thought prompting reported for similar tasks~\cite{wei2023chainofthought}. However, we argue that chain-of-thought reasoning is still helpful for explainability. 

Our game generation results show that being able to generate NPR Sunday Puzzle-style games is beyond the capabilities of current LLMs, even if they are capable of solving them. Future work could explore fine-tuning a model on our dataset rather than using few-shot learning. We hope that the PUZZLEQA dataset will aid future work in this area.

\section{Author Contributions}
All authors contributed to the writing of this paper.

\section{Acknowledgements}

We would like to thank our anonymous ICCC reviewers for their helpful feedback. We also extend our thanks to Eni Mustafaraj and Brian Tjaden for their comments on this work at various stages in its development, and to Arjun Guha for the initial idea of exploring the NPR Sunday Puzzle game.

\bibliographystyle{iccc}
\bibliography{iccc}

\end{document}